\documentclass{article}
\usepackage{ijcai22}

\usepackage{times}
\usepackage{soul}
\usepackage{url}
\usepackage[hidelinks]{hyperref}
\usepackage[utf8]{inputenc}
\usepackage[small]{caption}
\usepackage{graphicx}
\usepackage{amsmath}
\usepackage{amsthm}
\usepackage{booktabs}
\usepackage{algorithm}
\usepackage{algorithmic}
\usepackage{multirow,array}
\usepackage{amssymb}
\usepackage{pifont}
\usepackage{balance}
\usepackage{natbib}
\usepackage{diagbox}
\newcommand{\xmark}{\ding{55}}
\title{Legislator Representation Learning with Social Context and Expert Knowledge}

\author{
Shangbin Feng$^1$\and
Zhaoxuan Tan$^1$\and
Zilong Chen$^1$\and
Peisheng Yu$^1$\and \\
Qinghua Zheng$^1$\and
Xiaojun Chang$^2$\and
Minnan Luo$^1$
\affiliations
$^1$Xi'an Jiaotong University, Xi'an, Shaanxi, China\\
$^2$RMIT University, Melbourne, Victoria, Australia
\emails
\{wind\_binteng, tanzhaoxuan, luoyangczl, yps2000jj\}@stu.xjtu.edu.cn, \\
qhzheng@mail.xjtu.edu.cn,
cxj273@gmail.com,
minnluo@xjtu.edu.cn
}

\begin{document}

\maketitle

\begin{abstract}
Modeling the ideological perspectives of political actors is an essential task in computational political science with applications in many downstream tasks. Existing approaches are generally limited to textual data and voting records, while they neglect the rich social context and valuable expert knowledge for holistic evaluation. In this paper, we propose a representation learning framework of political actors that jointly leverages social context and expert knowledge. Specifically, we retrieve and extract factual statements about legislators to leverage social context information. We then construct a heterogeneous information network to incorporate social context and use relational graph neural networks to learn legislator representations. Finally, we train our model with three objectives to align representation learning with expert knowledge, model ideological stance consistency, and simulate the echo chamber phenomenon. Extensive experiments demonstrate that our learned representations successfully advance the state-of-the-art in three downstream tasks. Further analysis proves the correlation between learned legislator representations and various socio-political factors, as well as bearing out the necessity of social context and expert knowledge in modeling political actors.
\end{abstract}

\section{Introduction}
Modeling the perspectives of political actors has applications in various downstream tasks such as roll call vote prediction ~\citep{mou2021align} and political perspective detection ~\citep{feng2021knowledge}. Existing approaches generally focus on voting records or textual information of political actors to induce their stances. Ideal point model ~\citep{clinton2004statistical} is one of the most widely used approach for vote-based analysis, while later works enhance the ideal point model ~\citep{kraft2016embedding,gu2014topic,gerrish2011predicting} and yield promising results on the task of roll call vote prediction. For text-based methods, text analysis techniques are combined with textual information in social media posts ~\citep{li2019encoding}, Wikipedia pages ~\citep{feng2021knowledge}, legislative text ~\citep{mou2021align} and news articles ~\citep{li2021using} to enrich the perspective analysis process.

\begin{figure}
    \centering
    \includegraphics[width=1.0\linewidth]{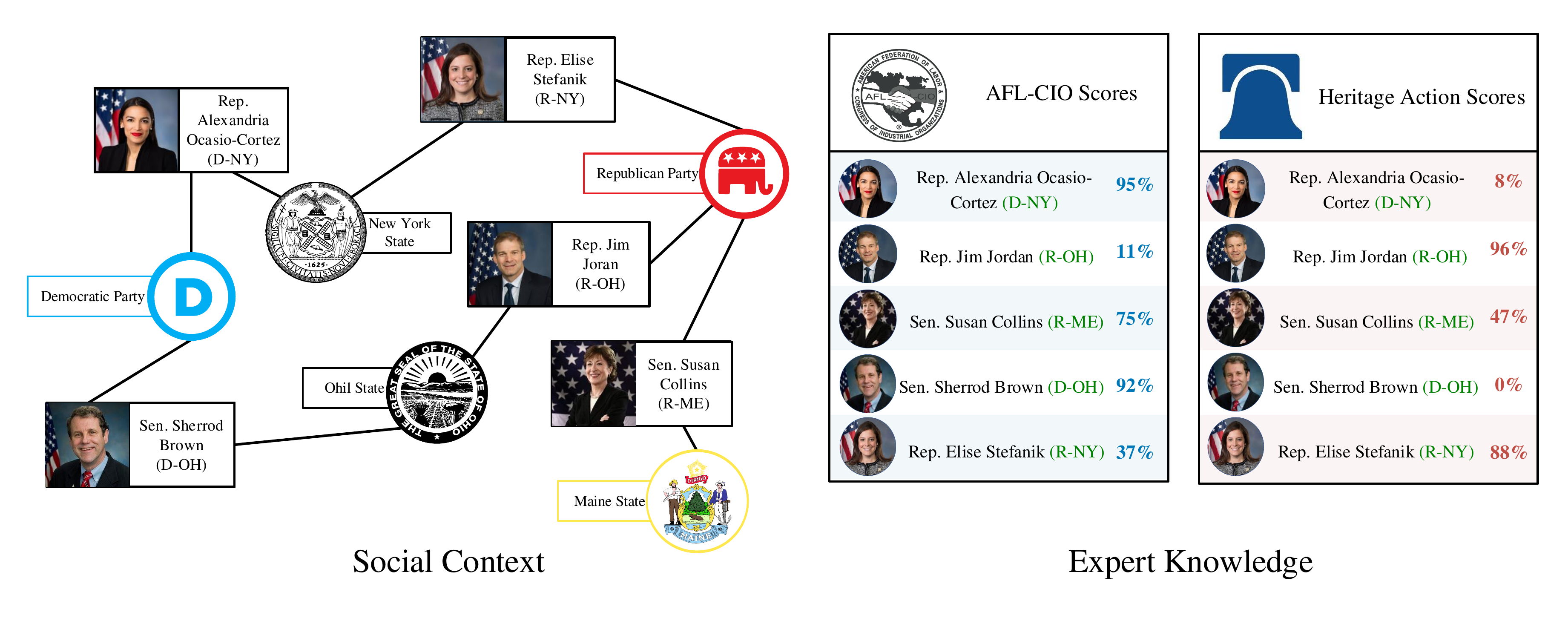}
    \caption{Social context information and expert knowledge from political think tanks to facilitate modeling different political actors.}
    \label{fig:teaser}
\end{figure}

However, existing methods fail to incorporate the rich social context and valuable expert knowledge. As illustrated in Figure \ref{fig:teaser}, social context information such as home state and party affiliation serves as background knowledge and helps connect different political actors \citep{yang2021joint}. These social context facts about political actors also differentiate them and indicate their ideological stances. In addition, expert knowledge from political think tanks provides valuable insights and helps to anchor the perspective analysis process. As a result, political actor representation learning should be guided by domain expertise to facilitate downstream tasks in computational political science. That being said, social context and expert knowledge should be incorporated in modeling legislators to ensure a holistic evaluation process.


\begin{figure*}
    \centering
    \includegraphics[width=0.8\linewidth]{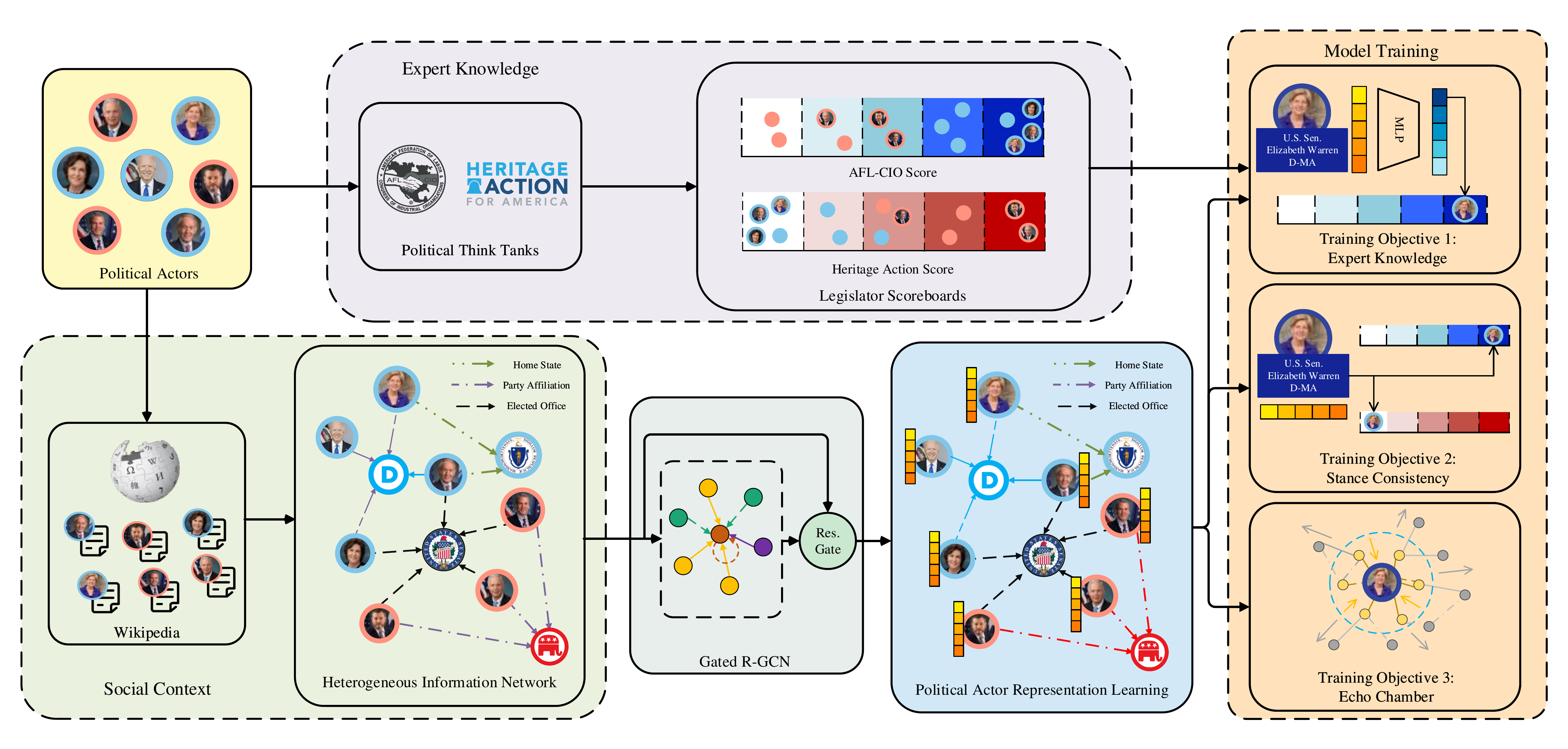}
    \caption{Overview of our proposed framework for political actor representation learning.}
    \label{fig:overview}
\end{figure*}

In light of these challenges, we propose a legislator representation learning framework that jointly leverages social context and expert knowledge. We firstly collect a dataset of political actors by retrieving and extracting social context information from their Wikipedia homepages and adapting expert knowledge from two political think tanks AFL-CIO\footnote{\url{https://aflcio.org/}} and Heritage Action\footnote{\url{https://heritageaction.com/}}. After that, we construct a heterogeneous information network to model social context information and adopt relational graph neural networks for representation learning. Finally, we train the framework with three training objectives to leverage expert knowledge, model social and political phenomena, and learn representations of political actors in the process. We apply our learned representations to three downstream tasks, namely political perspective detection, roll call vote prediction and entity stance prediction, to examine the effectiveness of our proposed approach. Our main contributions are summarized as follows:

\begin{itemize}
    \item To the best of our knowledge, this is the first work to jointly leverage social context and expert knowledge to learn representations of political actors.
    \item We propose a heterogeneous graph-based approach to learn legislator representations with three training objectives, which aligns social context with expert knowledge, ensures stance consistency, and models the echo chamber phenomenon in socio-economic systems.
    \item Extensive experiments demonstrate that our proposed approach outperforms the state-of-the-art in three related downstream tasks. Further studies suggest our learned legislator representations reflect various socio-political factors and prove the necessity of social context and expert knowledge in our proposed approach.
    
\end{itemize}



\section{Related Work}
The ideological perspectives of political actors play an essential role in their individual behaviour and adds up to influence the overall legislative process. Political scientists first explored to quantitatively model political actors based on their voting behaviour. Ideal point model~\citep{clinton2004statistical} is one of the earliest approach to leverage voting records to analyze their perspectives. It projects political actors and legislation onto one-dimensional spaces and measure distances. Many works later extended the ideal point model. \citet{gerrish2011predicting} leverages bill content to infer legislator perspectives. \citet{gu2014topic} introduces topic factorization to model voting behaviour on different issues to establish a fine-grained approach. \citet{kraft2016embedding} models legislators with multidimensional ideal vectors to analyze voting records.


In addition to voting, various forms of textual data such as speeches and public statements are also leveraged to model the perspectives of political actors. \citet{li2019encoding} proposes to analyze social media posts to better understand the stances of political content. \citet{feng2021knowledge} introduces textual information on Wikipedia and constructs a knowledge graph to facilitate perspective detection. \citet{mou2021align} proposes to leverage tweets, hashtags and legislative text to grasp the full picture of the political discourse. \citet{li2021using} focuses on analyzing the political perspectives of news articles and their mentioned political entities. In this paper, we explore to leverage social context and expert knowledge for legislator representation learning.


\section{Methodology}
Figure \ref{fig:overview} presents an overview of our political actor representation learning framework. We firstly collect data of political actors from Wikipedia and political think tanks. We then construct a heterogeneous information network to jointly model political actors, their social context and expert knowledge. After that, we learn graph representations with gated relational graph convolutional networks (gated R-GCN) and train the proposed framework with three different objectives.

\subsection{Data Collection}
We collect a datset about political actors in the United States that were active in the past decade while our proposed approach is applicable for all nations and time ranges. For \textbf{social context} information, we retrieve the list of senators and congresspersons from the 114th congress to the 117th congress.\footnote{https://www.congress.gov/} We then retrieve their Wikipedia pages\footnote{https://github.com/goldsmith/Wikipedia} and extract these named entities: presidents, senators, congresspersons, governors, states, political parties, supreme court justices, government institutions, and office terms (117th congress etc.). In this way, we obtain 1,069 social and political entities. Based these entities, we identify five types of relations: party affiliation, home state, political office, term in office, and appoint relationships. In this way, we obtain 9,248 heterogeneous edges. For \textbf{expert knowledge}, we make use of the legislator scoreboards at AFL-CIO and Heritage Action, two political think tanks that lie in opposite ends of the ideological spectrum. Specifically, we retrieve the scoreboard content and extract each legislator's score in each office term. In this way, we obtain 777 scores from AFL-CIO and 679 scores from Heritage Action. We consolidate the collected social context and expert knowledge to serve as the data set in our experiments, which is available as supplementary material.



\subsection{Graph Construction}
To better model the interactions between political entities and their shared social context, we propose to construct a heterogeneous information network (HIN) from the dataset. For initial node features, we use pre-trained RoBERTa \citep{liu2019roberta} to encode the first paragraph of Wikipedia homepage.

\subsubsection{Heterogeneous Nodes}
Based on the collected dataset, we select diversified entities that are essential factors in modeling the political process. Specifically, we use eight types of nodes to represent political actors and diversified social context entities.

\noindent \textit{\underline{N1: Office Terms}} We use four nodes to represent the 114th, 115th, 116th, 117th congress spanning from 2015 to 2021. We use these nodes to model the change in politics through time and could be similarly extended to other time periods.

\noindent \textit{\underline{N2: Legislators}} We retrieve senators and congresspersons from the 114th to the 117th congress and use one node to represent each distinct legislator.

\noindent \textit{\underline{N3: Presidents}} The presidency is the highest elected office in the United States. We use three nodes to represent President Biden, Trump and Obama to match with legislators.

\noindent \textit{\underline{N4: Governors}} State and local politics are also essential in analyzing the political process. We use one node to represent each distinct governor of 50 states.

\noindent \textit{\underline{N5: States}} The home state of political actors is often an important indicator and helps connect different individuals. We use one node to represent each state in the United States.

\noindent \textit{\underline{N6: Government Institutions}} We use four nodes to represent the white house, senate, house of representatives and supreme court. These nodes enable our constructed HIN to separate different political actors based on the office they hold.

\noindent \textit{\underline{N7: Supreme Court Justices}} Supreme court justices are nominated by presidents and approved by senators, which helps connect different types of political actors. We use one node to represent each supreme court justice.

\noindent \textit{\underline{N8: Political Parties}} We use two nodes to represent the two major political parties in the United States: the Republican Party and the Democratic Party.

\subsubsection{Heterogeneous Relations}
\label{subsubsec:relation}
Based on $N1$ to $N8$, we extract five types of informative interactions between entities to complete the HIN structure. Specifically, we use five types of heterogeneous relations to connect different nodes and construct our political actor HIN.

\noindent \textit{\underline{R1: Party Affiliation}} We connect political actors and their affiliated political party with $R1$:
\begin{equation}
R1 = (N2 \cup N3 \cup N4) \times N8
\end{equation}

\noindent \textit{\underline{R2: Home State}} We connect political actors with their home states with $R2$: 
\begin{equation}
R2 = (N2 \cup N3 \cup N4 \cup N7) \times N4
\end{equation}

\noindent \textit{\underline{R3: Hold Office}} We connect political actors with the political office they hold with $R3$:
\begin{equation}
R3 = (N2 \cup N3 \cup N4 \cup N7) \times N6
\end{equation}

\noindent \textit{\underline{R4: Time in Office}} If a political actor holds office during one of the time stamps in $N1$, we connect them with $R4$:
\begin{equation}
R4 = (N2 \cup N3 \cup N4 \cup N7) \times N1
\end{equation}

\noindent \textit{\underline{R5: Appoint}} Besides from being elected, certain political actors are appointed by others. We denote this relation with $R5$:
\begin{equation}
R5 = (N3 \times N7) \cup (N4 \times N2)
\end{equation}


\subsection{Representation Learning}
Since nodes represent political actors, we learn node-level representations with gated R-GCN to jointly leverage social context and external expert knowledge. Let $E = \{e_1, \cdot \cdot \cdot, e_n\}$ be $n$ entities and $v_i$ be the initial features of entity $e_i$. Let $R$ be the heterogeneous relation set and $N_r(e_i)$ be entity $e_i$'s neighborhood under relation type $r$. We firstly transform $v$ to serve as the input of graph neural networks,
\begin{equation}
    x_i^{(0)} = \phi(W_I \cdot v_i + b_I)
\end{equation}
where $\phi$ is leaky-relu, $W_I$ and $b_I$ are learnable parameters. We then propagate entity messages and aggregate them with gated R-GCN. For the $l$-th layer,
\begin{equation}
    u_i^{(l)} = \sum_{r \in R} \frac{1}{|N_r(e_i)|} \sum_{j \in N_r(e_i)}  f_r(x_j^{(l-1)}) + f_s(x_i^{(l-1)})
\end{equation}
where $f_s$ and $f_r$ are parameterized linear layers for self loops and edges of relation $r$, $u_i^{(l)}$ is the hidden representation for entity $e_i$ at layer $l$. We then calculate gate levels,
\begin{equation}
    g_i^{(l)} = \sigma(W_G \cdot [u_i^{(l)}, x_i^{(l-1)}] + b_G)
\end{equation}
where $W_G$ and $b_G$ are learnable parameters, $\sigma(\cdot)$ denotes the sigmoid function and $[\cdot , \cdot]$ denotes the concatenation operation. We then apply the gate mechanism to $u_i^{(l)}$ and $x_i^{(l-1)}$,
\begin{equation}
    x_i^{(l)} = tanh(u_i^{(l)}) \odot g_i^{(l)} + x_i^{(l-1)} \odot (1 - g_i^{(l)})
\end{equation}
where $\odot$ is the Hadamard product operation. After $L$ layer(s) of gated R-GCN, we obtain node representations $\{x^{(L)}_1,\cdot \cdot \cdot,x^{(L)}_n\}$ and the nodes representing political actors are extracted as learned representations.

\subsection{Model Training}
We propose to train our framework with a combination of supervised, self-supervised and unsupervised tasks, which aligns social context with expert knowledge, ensures stance consistency and simulates the echo chamber phenomenon. The overall loss function of our model is as follows:

\begin{equation}
    L = \lambda_1 L_1 + \lambda_2 L_2 + \lambda_3 L_3 + \lambda_4 \sum_{w \in \theta} w^2
\label{equ:total_loss}
\end{equation}

\noindent where $\lambda_i$ is the weight of loss $L_i$ and $\theta$ are all learnable parameters in the model. We then present the motivation and details of each loss function $L_1$, $L_2$ and $L_3$.

\begin{table}[]
    \centering
    \begin{tabular}{c|c c|c c}
         \toprule[1.5pt] \multirow{2}{*}{\textbf{Method}} & \multicolumn{2}{c|}{\textbf{SemEval}} & \multicolumn{2}{c}{\textbf{AllSides}} \\ 
         & \textbf{Acc} & \textbf{MaF} & \textbf{Acc} & \textbf{MaF} \\ \midrule[0.75pt]
         BERT & $84.03$ & $81.55$ & $81.55$ & $80.13$ \\
         MAN\_Glove & $81.58$ & $79.29$ & $78.29$ & $76.96$ \\
         MAN\_ELMO & $84.66$ & $83.09$ & $81.41$ & $80.44$ \\
         MAN\_Ensemble & $86.21$ & $84.33$ & $85.00$ & $84.25$ \\
         \citet{feng2021knowledge} & $89.56$ & $84.94$ & $86.02$ & $85.52$ \\
         \textbf{Ours} & $\textbf{91.30}$ & $\textbf{87.78}$ & $\textbf{86.81}$ & $\textbf{86.33}$ \\ 
         \bottomrule[1.5pt]
    \end{tabular}
    \caption{Adapting our learned representations for political perspective detection. Acc and MaF denote accuracy and macro F1-score.}
    \label{tab:downstream_perspective}
\end{table}

\subsubsection{Objective 1: Expert Knowledge}
The expert knowledge task aims to align the learned representations with external expert knowledge from political think tanks. We use the learned representations of political actors to predict their liberal and conservative stances, which are adapted from AFL-CIO and Heritage Action. Specifically,
\begin{equation}
\begin{aligned}
    l_i = softmax(W_L \cdot x_i^{(L)} + b_L) \\
    c_i = softmax(W_C \cdot x_i^{(L)} + b_C)
\end{aligned}
\end{equation}
where $l_i$ and $c_i$ are predicted stances towards liberal and conservative values, $W_L$, $b_L$, $W_C$ and $b_C$ are learnable parameters. Let $E_L$ and $E_C$ denote the training set of AFL-CIO and Heritage Action scores, $\hat{l_i}$ and $\hat{c_i}$ denote the ground-truth labels. We calculate the expert knowledge loss,

\begin{equation}
    L_1 = - \sum_{e_i \in E_L} \sum_{d=1}^D \hat{l_{id}}log(l_{id}) - \sum_{e_i \in E_C} \sum_{d=1}^D \hat{c_{id}}log(c_{id})
\end{equation}

$L1$ enables the framework to align learned representations with external expert knowledge from political think tanks.

\subsubsection{Objective 2: Stance Consistency}
The stance consistency task is motivated by the fact that liberalism and conservatism are opposite ideologies, thus individuals often take inversely correlated stances towards them. We firstly speculate entities' stance towards the opposite ideology by taking the opposite of the predicted stance,
\begin{equation}
    \Tilde{l_i} = \psi(D - argmax(c_i)), \ \ \Tilde{c_i} = \psi(D - argmax(l_i))
\end{equation}
where $\psi$ is the one-hot encoder, $argmax(\cdot)$ calculates the vector index with the largest value, $D$ is the number of stance labels, $\Tilde{l_i}$ and $\Tilde{c_i}$ are labels derived with stance consistency. We calculate the loss function $L_2$ measuring stance consistency,
\begin{equation}
    L_2 = - \sum_{e_i \in E} \ \sum_{d=1}^D \ (\Tilde{l_{id}} \ log(l_{id}) \ + \ \Tilde{c_{id}} \ log(c_{id}) )
\end{equation}

As a result, the loss function $L_2$ enables our approach to ensure stance consistency among political actors.

\subsubsection{Objective 3: Echo Chamber}
The echo chamber task is motivated by the echo chamber phenomenon, where social entities tend to reinforce their narratives by forming small and closely connected interaction circles. We simulate echo chambers by assuming that neighboring nodes in the HIN have similar representations while non-neighboring nodes have different representations. We firstly define the positive and negative neighborhood of entity $e_i$,
\begin{equation}
\begin{aligned}
P_{e_i} = \{e \ \ | \ \exists \ r \in R \ \ s.t. \ \ e \in N_r(e_i) \} \\
N_{e_i} = \{e \ \ | \ \forall \ r \in R \ \ s.t. \ \ e \notin N_r(e_i) \}
\end{aligned}
\end{equation}

We then calculate the echo chamber loss,
\begin{equation}
\begin{aligned}
    L_3 = - \sum_{e_i \in E} \sum_{e_j \in P_{e_i}} log(\sigma(x_i^{{(L)^T}} x_j^{(L)})) \\
    + Q \cdot \sum_{e_i \in E} \sum_{e_j \in N_{e_i}} log(\sigma(-x_i^{{(L)^T}} x_j^{(L)}))
\end{aligned}
\label{equ:unsupervised_loss}
\end{equation}
where $Q$ denotes the weight for negative samples. $L_3$ enables our framework to model the echo chamber phenomenon that is common in real-world socio-economic systems.

\begin{figure}[]
    \centering
    \includegraphics[width=0.85\linewidth]{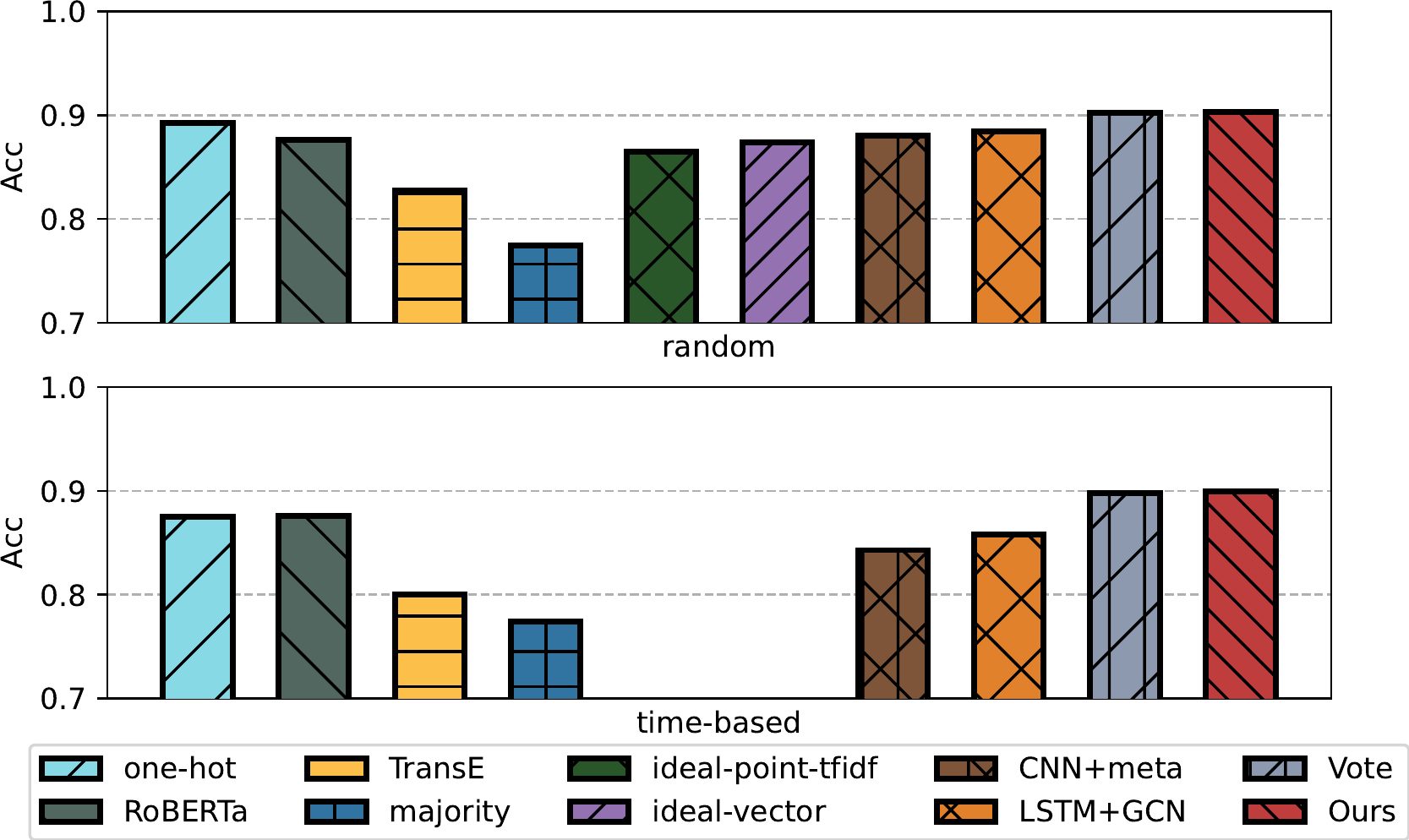}
    \caption{Adapting our learned representations of political actors for the downstream task of roll call vote prediction.}
    \label{fig:downstream_vote}
\end{figure}

\section{Experiments}
\subsection{Experiment Settings}
We train our representation learning framework with the proposed loss function $L$ in Equ (\ref{equ:total_loss}) and evaluate the learned representations of political actors on three downstream tasks: political perspective detection, roll call vote prediction, and entity stance prediction. We submit our data, implemented codes, hyperparameter settings and other experiment details as supplementary material to facilitate reproduction.

\begin{figure*}[]
    \centering
    \includegraphics[width=0.8\linewidth]{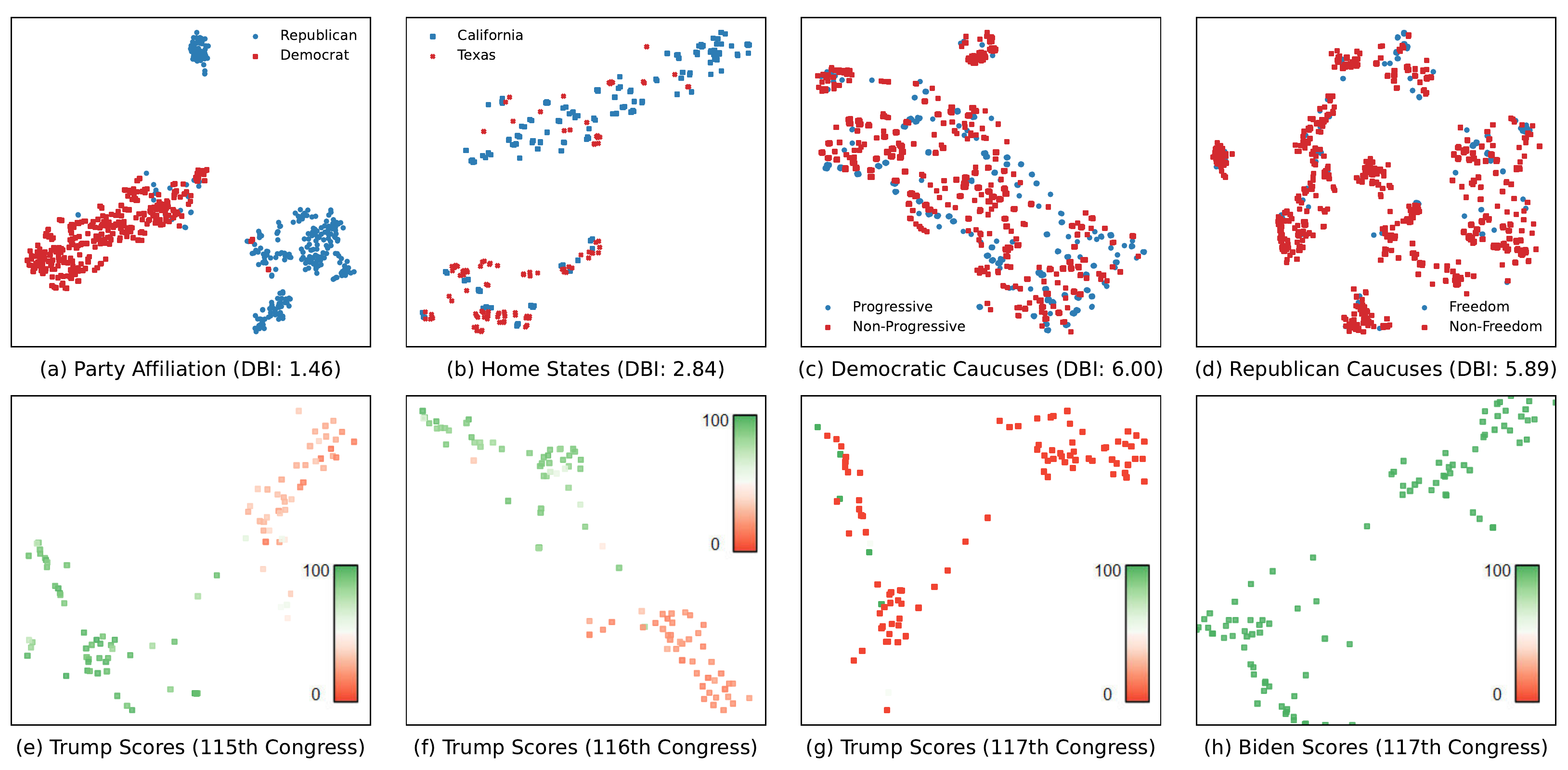}
    \caption{Using t-sne to visualize learned representations of political actors. DBI denotes the Davies-Bouldin Index.}
    \label{fig:replearn}
\end{figure*}

\subsubsection{Political Perspective Detection}
Political perspective detection aims to detect stances in text such as public statements and news articles, which generally mention many political actors to provide context and present arguments. \citet{feng2021knowledge} proposes to leverage TransE \citep{TransE} to learn representations of political actors and augment the argument mining process. To examine the quality of our representation learning, we replace TransE in \citet{feng2021knowledge} with our learned representations and conduct political perspective detection. We adopt the same datasets \citep{SemEval,li2019encoding} and settings so that the results are directly comparable.

\subsubsection{Roll Call Vote Prediction}
Roll call vote prediction aims to predict the voting behavior of legislators given different pieces of legislation, which is essential in the political process and involves analyzing political actors. To examine our representations' effectiveness in vote prediction, we concatenate our learned representations with RoBERTa \citep{liu2019roberta} encoding of legislation texts and conduct vote prediction with two fully connected layers. We adopt the same dataset and experiment settings as \citet{mou2021align} so that the results are directly comparable.

\subsubsection{Expert Knowledge Prediction}
To the best of our knowledge, this is the first work to leverage expert knowledge from political think tanks in modeling political perspectives. To examine whether our learned representations correlate well with expert knowledge, we compare with various text~\citep{pedregosa2011scikit,MEANbiasfeature,pennington2014glove,liu2019roberta,beltagy2020longformer} and graph~\citep{GCN,GAT,SAGE,TransformerConv,ResGatedGraphConv} analysis baselines on the expert knowledge task.

\begin{table}[]
    \centering
    \resizebox{\linewidth}{!}{
    \begin{tabular}{c|c c|c c c}
         \toprule[1.5pt] \textbf{Method} & \textbf{Text} & \textbf{Graph} & \textbf{Acc} & \textbf{MaF} & \textbf{MiF} \\ \midrule[0.75pt]
         Linear BoW & \checkmark &  & $68.49$&	$40.00$&	$68.53$ \\
         Bias Features & \checkmark & & $47.26$&	$20.08$&	$47.10$ \\
         Glove & \checkmark & & $52.05$&	$26.94$&	$52.01$ \\
         RoBERTa & \checkmark & & $71.92$&	$49.70$&	$71.87$ \\
         LongFormer & \checkmark & & $68.49$&	$42.27$&	$68.56$ \\ \midrule[0.75pt]
         GCN & \checkmark & \checkmark & $74.66$&	$54.16$&	$74.46$ \\
         GAT & \checkmark & \checkmark & $78.08$&	$55.82$&	$78.17$ \\
         GraphSAGE & \checkmark & \checkmark & $75.34$&	$51.39$&	$75.43$ \\
         TransformerConv & \checkmark & \checkmark & $77.40$&	$55.63$&	$77.48$ \\
         ResGatedConv & \checkmark & \checkmark & $76.03$&	$54.31$&	$75.97$ \\ \midrule[0.75pt]
         \textbf{Ours} & \checkmark & \checkmark &  $\textbf{80.82}$&	$\textbf{60.37}$&	$\textbf{80.89}$ \\ \bottomrule[1.5pt]
    \end{tabular}
    }
    \caption{Our model's performance on the expert knowledge prediction task compared to various text and graph analysis baselines. Acc, MaF and MiF denote accuracy, macro and micro averaged F1-score.}
    \label{tab:big}
\end{table}

\subsection{Results}
\label{subsec:big}

\subsubsection{Political Perspective Detection}
Table \ref{tab:downstream_perspective} presents model performance on two political perspective detection datasets. We run our approach five times and report the average performance. It is demonstrated that our approach consistently outperforms all state-of-the-art approaches on both benchmarks, which indicates that our learned representations of political actors could serve as external knowledge in the task of political perspective detection to help augment the argument mining process.

\subsubsection{Roll Call Vote Prediction}
Figure \ref{fig:downstream_vote} presents model performance on roll call vote prediction. We run ours five times and report the average results. It is demonstrated that our approach outperforms existing baselines, which suggests that our representation learning could serve as high-quality features to analyze voting behaviour.

\subsubsection{Expert Knowledge Prediction}
Table \ref{tab:big} presents model performance on the expert knowledge task. We run \textbf{Ours} five times and report the average performance. It is demonstrated that our framework consistently achieves the best performance compared to various graph and analysis baselines, suggesting our learned representations successfully reflect expert knowledge from political think tanks. Apart from that, graph-based models generally outperform text-based methods, which suggests that the constructed HIN is essential to model performance.

\subsection{Representation Learning Study}
Section \ref{subsec:big} demonstrates that our framework successfully learns political actor representations that improve performance in three downstream tasks. Besides, high-quality representations of political actors should differentiate them with regard to their social and political information. To examine whether we have achieved this end, we adopt t-sne \citep{maaten_visualizing_2008} to illustrate our learned representations of political actors in Figure \ref{fig:replearn}.

\subsubsection{Social Context}
Figure \ref{fig:replearn} (a) and (b) illustrate the correlation between learned representations and social context. The DBI scores \citep{DBIscore} quantitatively indicate great collocation among different social context groups.

\subsubsection{Congressional Caucuses} Figure \ref{fig:replearn} (c) and (d) demonstrate the correlation between learned representations and major congressional caucuses in both parties. Both the illustration and the DBI scores suggest little collocation among caucuses, which could be attributed to the fact that inter-party differences outweigh intra-party differences in contemporary U.S. politics.

\begin{figure}
    \centering
    \includegraphics[width=0.83\linewidth]{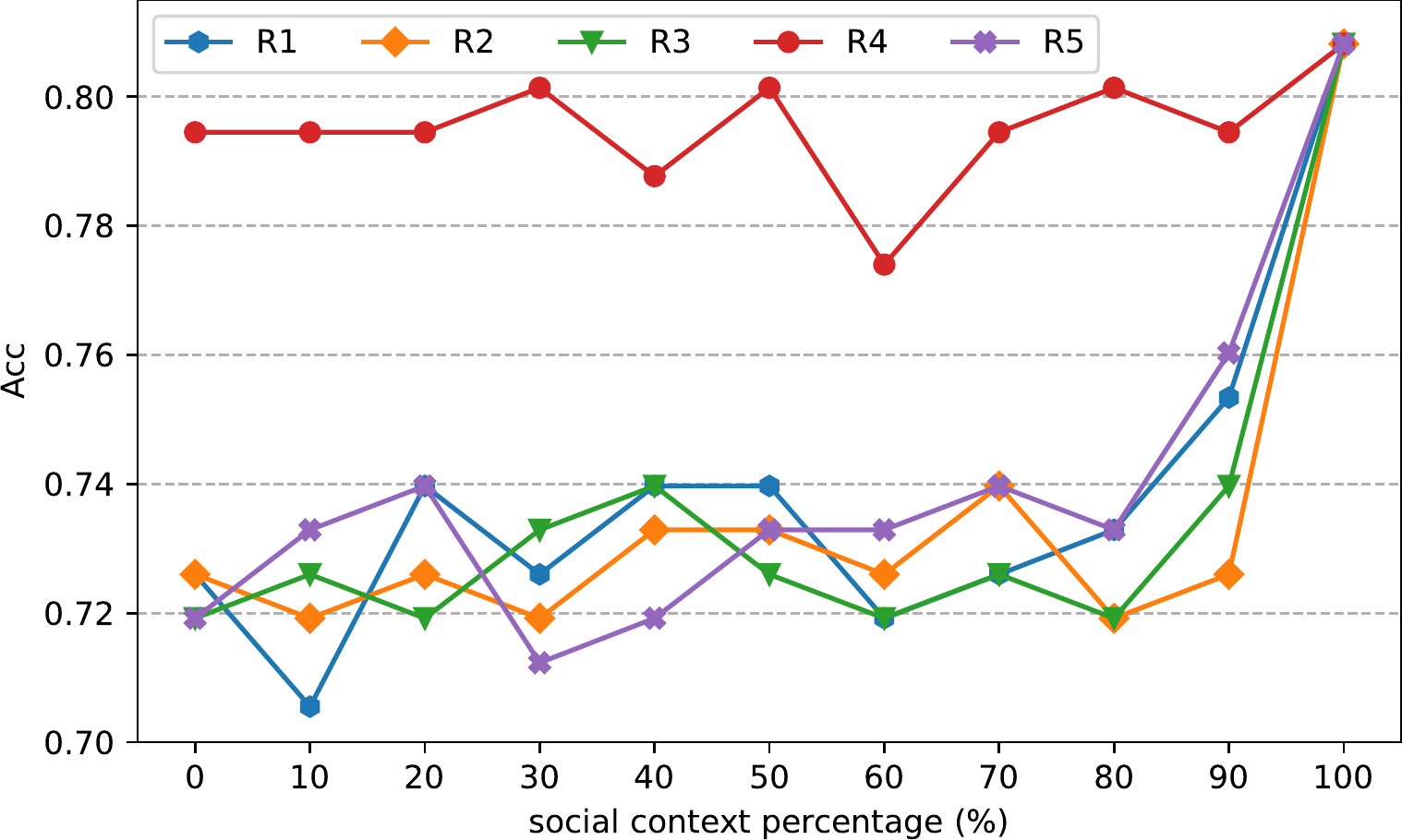}
    \caption{Ablation study of social context information. Definition of five types of relations $R1$ to $R5$ follows that in Section \ref{subsubsec:relation}.}
    \label{fig:ablation_social_context}
\end{figure}

\subsubsection{Voting Records} Figure \ref{fig:replearn} (e), (f), (g) and (h) illustrate how legislators vote with or against the sitting president. We retrieve this information from FiveThirtyEight\footnote{\url{https://fivethirtyeight.com/}} and illustrate voting records with color gradients. As a result, our learned representations of legislators in the 115th and 116th congress correlate well with their voting records, while the 117th congress might have not hold enough votes for an accurate categorization.

\subsection{Ablation Study}
Our framework aims to learn representations of political actors with the help of social context and expert knowledge as well as three training objectives. We conduct ablation study to examine their effect in the representation learning process and report performance on the expert knowledge prediction.

\subsubsection{Social Context}
We use five types of heterogeneous relations $R1$ to $R5$ to connect different entities based on their social context. We gradually remove five types of social context edges in the constructed HIN and report model performance in Figure \ref{fig:ablation_social_context}. It is illustrated that all relations but $R4$ (Time in Office) significantly contributes to the overall performance. Besides, Figure \ref{fig:ablation_social_context} illustrates a great gap between $90\%$ and $100\%$ edges, suggesting the importance of a complete HIN structure.


\subsubsection{Expert Knowledge}
\begin{figure}
    \centering
    \includegraphics[width=0.8\linewidth]{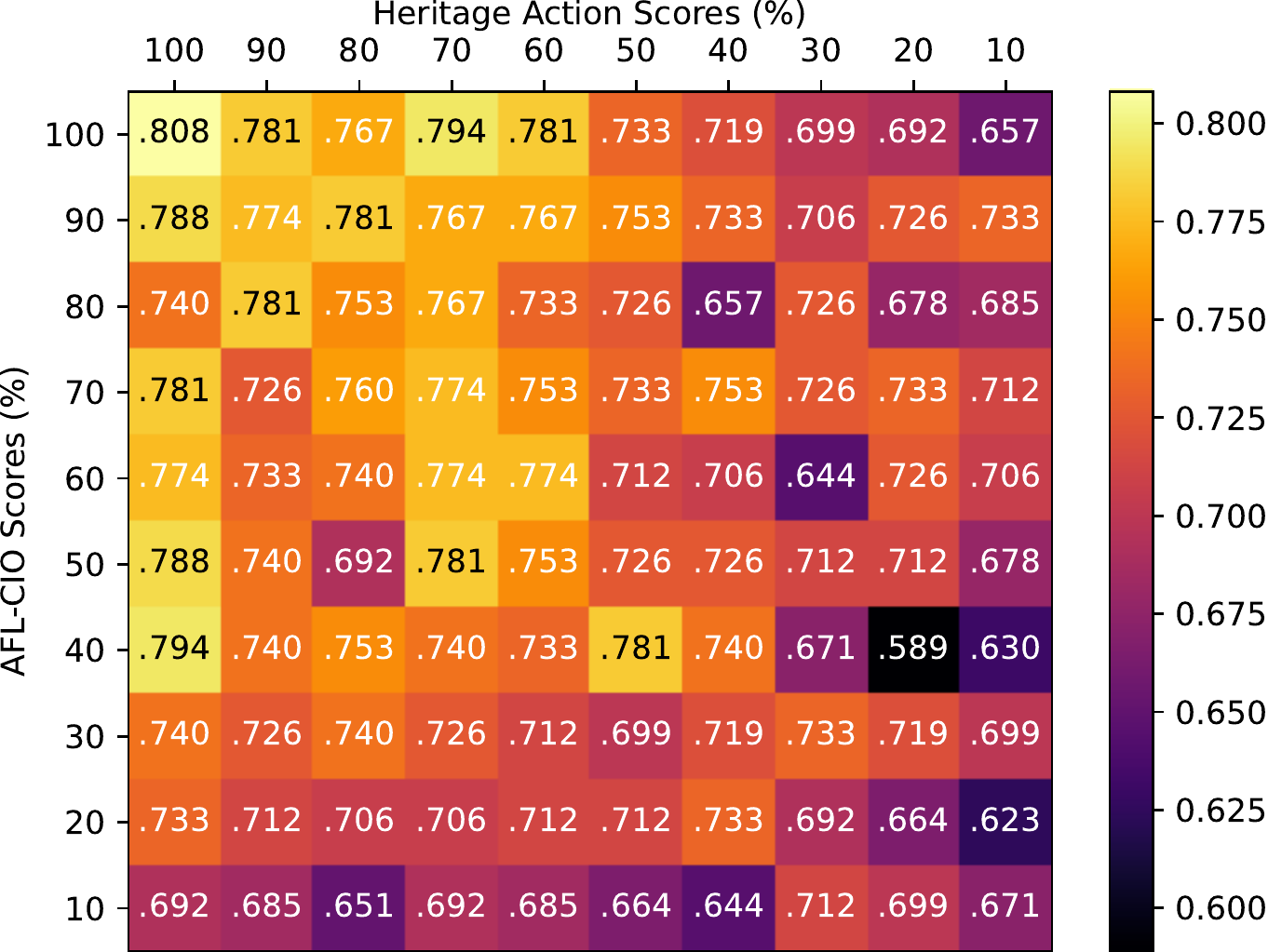}
    \caption{Ablation study of expert knowledge. We report accuracy on the training objective 1 for different expert knowledge settings.}
    \label{fig:ablation_expert_knowledge}
\end{figure}

We learn legislator representations with the help of two political think tanks: AFL-CIO and Heritage Action. We retrieve their evaluation of political actors amd construct a supervised task to train our framework. To examine the effect of these expert knowledge in our proposed approach, we gradually remove expert knowledge labels in $L_1$ and report model performance in Figure \ref{fig:ablation_expert_knowledge}. It is illustrated that our performance drops with partial expert knowledge from either AFL-CIO or Heritage Action. As a result, expert knowledge is essential in our model's representation learning process.

\subsubsection{Training Objectives}
We propose to train our framework with three objectives. To examine their effect, we train our method with different combinations of $L_1$, $L_2$ and $L_3$ and report performance in Table \ref{tab:ablation_training_objectives}. Our model performs best with all three training objectives, proving the effectiveness of our loss function design.

\begin{table}[]
    \centering
    \begin{tabular}{c|c c c}
         \toprule[1.5pt] \textbf{Loss Function(s)} & \textbf{Acc} & \textbf{MaF} & \textbf{MiF} \\ \midrule[0.75pt]
         $L_1$ only & $78.08$ & $56.43$ & $78.17$ \\
         $L_1$ and $L_2$ & $79.45$ & $55.97$ & $79.52$ \\
         $L_1$ and $L_3$ & $76.03$ & $51.91$ & $76.11$ \\
         $L_1$, $L_2$, and $L_3$ & $\textbf{80.82}$ & $\textbf{60.37}$ & $\textbf{80.89}$ \\ \bottomrule[1.5pt]
    \end{tabular}
    \caption{Ablation study of three training objectives.}
    \label{tab:ablation_training_objectives}
\end{table}


\section{Conclusion}
In this paper, we present a framework to learn representations of political actors with social context and expert knowledge. We retrieve context information from Wikipedia and expert knowledge from political think tanks, construct a HIN to model political actors and learn representations with gated R-GCNs and multiple training objectives. Extensive experiments demonstrate that our approach advances the state-of-the-art on three downstream tasks. Further experiments also prove the quality of learned representations and the necessity of social context and expert knowledge in our approach.


\bibliographystyle{named}
\balance
\bibliography{ijcai22}

\begin{thebibliography}{}

\bibitem[\protect\citeauthoryear{Beltagy \bgroup \em et al.\egroup
  }{2020}]{beltagy2020longformer}
Iz~Beltagy, Matthew~E Peters, and Arman Cohan.
\newblock Longformer: The long-document transformer.
\newblock {\em arXiv preprint arXiv:2004.05150}, 2020.

\bibitem[\protect\citeauthoryear{Bordes \bgroup \em et al.\egroup
  }{2013}]{TransE}
Antoine Bordes, Nicolas Usunier, Alberto Garcia-Duran, Jason Weston, and Oksana
  Yakhnenko.
\newblock Translating embeddings for modeling multi-relational data.
\newblock {\em Advances in neural information processing systems}, 26, 2013.

\bibitem[\protect\citeauthoryear{Bresson and Laurent}{2017}]{ResGatedGraphConv}
Xavier Bresson and Thomas Laurent.
\newblock Residual gated graph convnets.
\newblock {\em arXiv preprint arXiv:1711.07553}, 2017.

\bibitem[\protect\citeauthoryear{Clinton \bgroup \em et al.\egroup
  }{2004}]{clinton2004statistical}
Joshua Clinton, Simon Jackman, and Douglas Rivers.
\newblock The statistical analysis of roll call data.
\newblock {\em American Political Science Review}, 98(2):355--370, 2004.

\bibitem[\protect\citeauthoryear{Davies and Bouldin}{1979}]{DBIscore}
David~L Davies and Donald~W Bouldin.
\newblock A cluster separation measure.
\newblock {\em IEEE transactions on pattern analysis and machine intelligence},
  (2):224--227, 1979.

\bibitem[\protect\citeauthoryear{Falcon}{2019}]{pytorchlightning}
et~al. Falcon, WA.
\newblock Pytorch lightning.
\newblock {\em GitHub. Note:
  https://github.com/PyTorchLightning/pytorch-lightning}, 3, 2019.

\bibitem[\protect\citeauthoryear{Feng \bgroup \em et al.\egroup
  }{2021}]{feng2021knowledge}
Shangbin Feng, Minnan Luo, Zilong Chen, Qingyao Li, Xiaojun Chang, and Qinghua
  Zheng.
\newblock Knowledge graph augmented political perspective detection in news
  media.
\newblock {\em arXiv preprint arXiv:2108.03861}, 2021.

\bibitem[\protect\citeauthoryear{Fey and Lenssen}{2019}]{torchgeometric}
Matthias Fey and Jan~Eric Lenssen.
\newblock Fast graph representation learning with pytorch geometric.
\newblock {\em arXiv preprint arXiv:1903.02428}, 2019.

\bibitem[\protect\citeauthoryear{Gerrish and
  Blei}{2011}]{gerrish2011predicting}
Sean~M Gerrish and David~M Blei.
\newblock Predicting legislative roll calls from text.
\newblock In {\em Proceedings of the 28th International Conference on Machine
  Learning, ICML 2011}, 2011.

\bibitem[\protect\citeauthoryear{Gu \bgroup \em et al.\egroup
  }{2014}]{gu2014topic}
Yupeng Gu, Yizhou Sun, Ning Jiang, Bingyu Wang, and Ting Chen.
\newblock Topic-factorized ideal point estimation model for legislative voting
  network.
\newblock In {\em Proceedings of the 20th ACM SIGKDD international conference
  on Knowledge discovery and data mining}, pages 183--192, 2014.

\bibitem[\protect\citeauthoryear{Hamilton \bgroup \em et al.\egroup
  }{2017}]{SAGE}
William~L Hamilton, Rex Ying, and Jure Leskovec.
\newblock Inductive representation learning on large graphs.
\newblock In {\em Proceedings of the 31st International Conference on Neural
  Information Processing Systems}, pages 1025--1035, 2017.

\bibitem[\protect\citeauthoryear{Horne \bgroup \em et al.\egroup
  }{2018}]{MEANbiasfeature}
Benjamin~D Horne, Sara Khedr, and Sibel Adali.
\newblock Sampling the news producers: A large news and feature data set for
  the study of the complex media landscape.
\newblock In {\em Twelfth International AAAI Conference on Web and Social
  Media}, 2018.

\bibitem[\protect\citeauthoryear{Kiesel \bgroup \em et al.\egroup
  }{2019}]{SemEval}
Johannes Kiesel, Maria Mestre, Rishabh Shukla, Emmanuel Vincent, Payam Adineh,
  David Corney, Benno Stein, and Martin Potthast.
\newblock Semeval-2019 task 4: Hyperpartisan news detection.
\newblock In {\em Proceedings of the 13th International Workshop on Semantic
  Evaluation}, pages 829--839, 2019.

\bibitem[\protect\citeauthoryear{Kipf and Welling}{2016}]{GCN}
Thomas~N Kipf and Max Welling.
\newblock Semi-supervised classification with graph convolutional networks.
\newblock {\em arXiv preprint arXiv:1609.02907}, 2016.

\bibitem[\protect\citeauthoryear{Kraft \bgroup \em et al.\egroup
  }{2016}]{kraft2016embedding}
Peter Kraft, Hirsh Jain, and Alexander~M Rush.
\newblock An embedding model for predicting roll-call votes.
\newblock In {\em Proceedings of the 2016 conference on empirical methods in
  natural language processing}, pages 2066--2070, 2016.

\bibitem[\protect\citeauthoryear{Li and Goldwasser}{2019}]{li2019encoding}
Chang Li and Dan Goldwasser.
\newblock Encoding social information with graph convolutional networks
  forpolitical perspective detection in news media.
\newblock In {\em Proceedings of the 57th Annual Meeting of the Association for
  Computational Linguistics}, pages 2594--2604, 2019.

\bibitem[\protect\citeauthoryear{Li and Goldwasser}{2021}]{li2021using}
Chang Li and Dan Goldwasser.
\newblock Using social and linguistic information to adapt pretrained
  representations for political perspective identification.
\newblock In {\em Findings of the Association for Computational Linguistics:
  ACL-IJCNLP 2021}, pages 4569--4579, 2021.

\bibitem[\protect\citeauthoryear{Liu \bgroup \em et al.\egroup
  }{2019}]{liu2019roberta}
Yinhan Liu, Myle Ott, Naman Goyal, Jingfei Du, Mandar Joshi, Danqi Chen, Omer
  Levy, Mike Lewis, Luke Zettlemoyer, and Veselin Stoyanov.
\newblock Roberta: A robustly optimized bert pretraining approach.
\newblock {\em arXiv preprint arXiv:1907.11692}, 2019.

\bibitem[\protect\citeauthoryear{Maaten and
  Hinton}{2008}]{maaten_visualizing_2008}
Laurens van~der Maaten and Geoffrey Hinton.
\newblock Visualizing {Data} using t-{SNE}.
\newblock {\em Journal of Machine Learning Research}, 9(86):2579--2605, 2008.

\bibitem[\protect\citeauthoryear{Mou \bgroup \em et al.\egroup
  }{2021}]{mou2021align}
Xinyi Mou, Zhongyu Wei, Lei Chen, Shangyi Ning, Yancheng He, Changjian Jiang,
  and Xuan-Jing Huang.
\newblock Align voting behavior with public statements for legislator
  representation learning.
\newblock In {\em Proceedings of the 59th Annual Meeting of the Association for
  Computational Linguistics and the 11th International Joint Conference on
  Natural Language Processing (Volume 1: Long Papers)}, pages 1236--1246, 2021.

\bibitem[\protect\citeauthoryear{Paszke \bgroup \em et al.\egroup
  }{2019}]{paszke2019pytorch}
Adam Paszke, Sam Gross, Francisco Massa, Adam Lerer, James Bradbury, Gregory
  Chanan, Trevor Killeen, Zeming Lin, Natalia Gimelshein, Luca Antiga, et~al.
\newblock Pytorch: An imperative style, high-performance deep learning library.
\newblock {\em Advances in neural information processing systems},
  32:8026--8037, 2019.

\bibitem[\protect\citeauthoryear{Pedregosa \bgroup \em et al.\egroup
  }{2011}]{pedregosa2011scikit}
Fabian Pedregosa, Ga{\"e}l Varoquaux, Alexandre Gramfort, Vincent Michel,
  Bertrand Thirion, Olivier Grisel, Mathieu Blondel, Peter Prettenhofer, Ron
  Weiss, Vincent Dubourg, et~al.
\newblock Scikit-learn: Machine learning in python.
\newblock {\em the Journal of machine Learning research}, 12:2825--2830, 2011.

\bibitem[\protect\citeauthoryear{Pennington \bgroup \em et al.\egroup
  }{2014}]{pennington2014glove}
Jeffrey Pennington, Richard Socher, and Christopher~D Manning.
\newblock Glove: Global vectors for word representation.
\newblock In {\em Proceedings of the 2014 conference on empirical methods in
  natural language processing (EMNLP)}, pages 1532--1543, 2014.

\bibitem[\protect\citeauthoryear{Shi \bgroup \em et al.\egroup
  }{2020}]{TransformerConv}
Yunsheng Shi, Zhengjie Huang, Wenjin Wang, Hui Zhong, Shikun Feng, and Yu~Sun.
\newblock Masked label prediction: Unified message passing model for
  semi-supervised classification.
\newblock {\em arXiv preprint arXiv:2009.03509}, 2020.

\bibitem[\protect\citeauthoryear{Veli{\v{c}}kovi{\'c} \bgroup \em et al.\egroup
  }{2017}]{GAT}
Petar Veli{\v{c}}kovi{\'c}, Guillem Cucurull, Arantxa Casanova, Adriana Romero,
  Pietro Lio, and Yoshua Bengio.
\newblock Graph attention networks.
\newblock {\em arXiv preprint arXiv:1710.10903}, 2017.

\bibitem[\protect\citeauthoryear{Wolf \bgroup \em et al.\egroup
  }{2020}]{wolf-etal-2020-transformers}
Thomas Wolf, Lysandre Debut, Victor Sanh, Julien Chaumond, Clement Delangue,
  Anthony Moi, Pierric Cistac, Tim Rault, Rémi Louf, Morgan Funtowicz, Joe
  Davison, Sam Shleifer, Patrick von Platen, Clara Ma, Yacine Jernite, Julien
  Plu, Canwen Xu, Teven~Le Scao, Sylvain Gugger, Mariama Drame, Quentin Lhoest,
  and Alexander~M. Rush.
\newblock Transformers: State-of-the-art natural language processing.
\newblock In {\em Proceedings of the 2020 Conference on Empirical Methods in
  Natural Language Processing: System Demonstrations}, pages 38--45, Online,
  October 2020. Association for Computational Linguistics.

\bibitem[\protect\citeauthoryear{Yang \bgroup \em et al.\egroup
  }{2021}]{yang2021joint}
Yuqiao Yang, Xiaoqiang Lin, Geng Lin, Zengfeng Huang, Changjian Jiang, and
  Zhongyu Wei.
\newblock Joint representation learning of legislator and legislation for roll
  call prediction.
\newblock In {\em Proceedings of the Twenty-Ninth International Conference on
  International Joint Conferences on Artificial Intelligence}, pages
  1424--1430, 2021.

\end{thebibliography}


\appendix
\section{Supplementary Studies}
To better understand the effectiveness of our propose framework, we present additional studies that do not fit in the main paper. We report model performance on the expert knowledge prediction task in supplementary studies.

\subsection{Graph Learning Study}
We construct a HIN to model social context and adopt gated R-GCNs for representation learning, thus we further study the effect of these graph-related elements.


\subsubsection{GNN Operator}
We adopt five heterogeneous relations $R1$ to $R5$ to connect eight types of entities $N1$ to $N8$, thus our constructed graph is heterogeneous. To examine whether the graph heterogeneity contributes to model performance, we substitute gated R-GCNs with homogeneous GNNs such as GCN, GAT and GraphSAGE. Table \ref{tab:gnn_operator} shows that our model performs best with gated R-GCNs, proving the necessity of heterogeneous relations to represent diversified social context.

\subsubsection{GNN Layers}
Two or three GNN layers are typically adopted for node-level representation learning in similar tasks. To examine the effect of GNN layers in our proposed approach, we learn node representations with one to five layers of gated R-GCNs. Table \ref{tab:gnn_layer} illustrates that two layers of gated R-GCNs perform best, thus we adhere to this setting in other experiments.

\begin{figure}[t]
    \centering
    \includegraphics[width=0.8\linewidth]{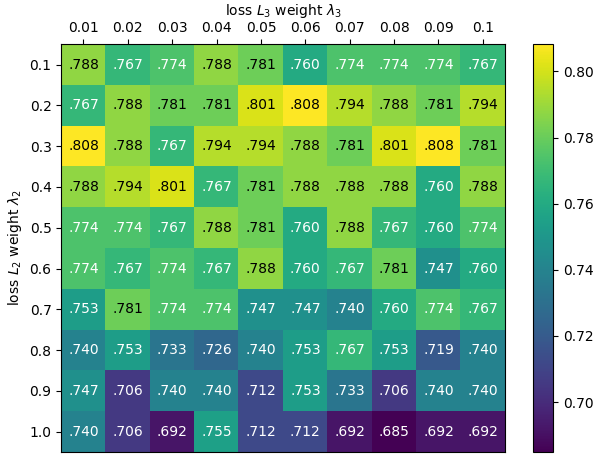}
    \caption{Model performance with different loss weights.}
    \label{fig:loss_weight}
\end{figure}

\begin{table}[]
    \centering
    \begin{tabular}{c c|c c c}
         \toprule[1.5pt] \textbf{GNN operator} & \textbf{Het.} & \textbf{Acc} & \textbf{MaF} & \textbf{MiF}  \\ \midrule[0.75pt]
         GCN & \xmark & 	$76.03$ &	$58.31$ &	$78.08$ \\
         GAT & \xmark &	$77.40$ &	$59.01$ &	$78.85$ \\
         SAGE & \xmark & $78.77$ &	$58.04$ & 	$78.81$ \\
         R-GCN & \checkmark & $78.08$ &	$55.61$&	$78.15$ \\
         Gated R-GCN & \checkmark & $\textbf{80.82}$ & $\textbf{60.37}$ & $\textbf{80.89}$ \\ \bottomrule[1.5pt]
    \end{tabular}
    \caption{Model performance with different GNN operators. Our approach adopts gated R-GCN and achieves the best performance. Het. denotes whether the method supports heterogeneous graphs.}
    \label{tab:gnn_operator}
\end{table}

\begin{table}[t]
    \centering
    \begin{tabular}{c | c | c | c | c | c}
         \toprule[1.5pt]
         $\textbf{L}$ & 0 & 1 & 2 & 3 & 4 \\
         \midrule[0.75pt]
        \textbf{Acc} & 69.18 & 74.66 & \textbf{80.82} & 78.08 & 78.08 \\ 
        \textbf{MaF} & 42.00 & 52.87 & \textbf{60.37} & 57.66 & 62.33\\ 
        \textbf{MiF} & 69.26 & 74.74 & \textbf{80.89} & 76.80 & 78.08\\ 
        \bottomrule[1.5pt]
    \end{tabular}
    \caption{Model performance with one to five gated R-GCN layers.}
    \label{tab:gnn_layer}
\end{table}

\begin{figure}[t]
    \centering
    \includegraphics[width=0.8\linewidth]{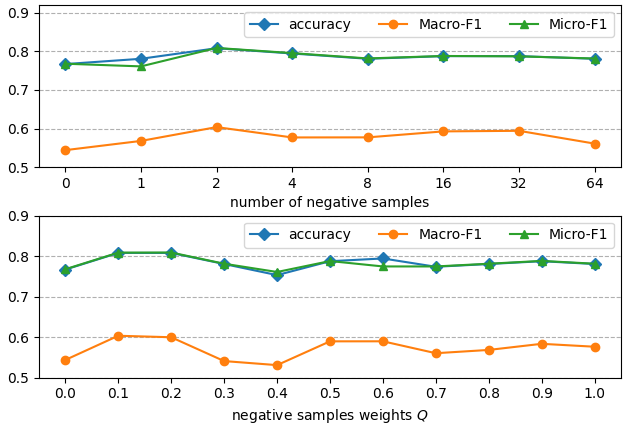}
    \caption{Model performance with different number of negative samples and weight $Q$.}
    \label{fig:negative_sample}
\end{figure}

\subsection{Training Objective Study}
We propose to train the representation learning framework with three objectives: expert knowledge, stance consistency, and the echo chamber phenomenon. We further study the effect of these training objectives in model performance.

\subsubsection{Loss Weight}
We fix $\lambda_1 = 1$ and $\lambda_4 = 10^{-5}$, present model performance under different settings of loss weights for auxiliary tasks $\lambda_2$ and $\lambda_3$ in Figure \ref{fig:loss_weight}. It is illustrated that $0.2 \leq \lambda_2 \leq 0.3$ and $0.01 \leq \lambda_3 \leq 0.1$ would generally lead to an effective balance of differently supervised loss functions.

\subsubsection{Negative Sample}
For the echo chamber objective, we randomly select nodes from $N_{e_i}$ to serve as negative samples and balance $L_3$ with $Q$. We study the effect of negative sample amount and their weight $Q$ in the model's performance and present them in Figure \ref{fig:negative_sample}. It is illustrated that 2 negative samples and $Q = 0.1$ or $0.2$ performs best, which indicates that the echo chamber objective contributes to overall performance, while too many negative samples and large $Q$s might overshadow the overall representation learning process.

\section{Experiment Details}
In this section, we provide additional details to facilitate reproducing our results and findings. We submit data and code as supplementary material and commit to make them publicly available upon acceptance.

\subsection{Hyperparameters}
We present the hyperparameter settings of our proposed approach in Table \ref{tab:hyperparameter}. We follow these settings throughout the paper unless stated otherwise.

\subsection{Implementation}
We use pytorch~\citep{paszke2019pytorch}, pytorch lightning~\citep{pytorchlightning}, torch geometric~\citep{torchgeometric} and the transformers library~\citep{wolf-etal-2020-transformers} for an efficient implementation of our proposed framework. All implemented codes are available as supplementary material.

\subsection{Experiment Details}

\subsubsection{Political Perspective Detection}
We replace TransE in \citet{feng2021knowledge} with our learned representations of political actors. We use the GRGCN setting in \citet{feng2021knowledge} as backbone. We maintain the same evaluation settings to ensure a fair comparison and highlight the effectiveness of our learned representations.

\subsubsection{Roll Call Vote Prediction}
We make our best effort to maintain the same experiment settings as \citet{mou2021align} while their might be minor differences. For \textit{random}, we conduct roll call vote prediction for legislators in the 114th and 115th congress and average the results. We follow the same 6:2:2 split for each setting. For \textit{time-based}, we use the 114th congress as training and validation set and the 115th congress as test set.

\subsubsection{Expert Knowledge Prediction}
We collect expert knowledge about legislators from two political think tanks, which assigns a continuous score $s$ from 0 to 1 to indicate how liberal or conservative a legislator is. We construct a classification task from expert knowledge by creating five discrete labels: strongly favor ($0.9 \le s \leq 1$), favor ($0.75 \le s \leq 0.9$), neutral ($0.25 \le s \leq 0.75$), oppose ($0.1 \le s \leq 0.25$), and strongly oppose ($0 \leq s \leq 0.1$). In this way, we adapt from expert knowledge to derive liberal and conservative labels for legislators. We use 7:2:1 to partition them into training, validation and test sets. We calculate evaluation metrics on the liberal and conservative set separately, and present the harmonic mean of two sets. In this way, the presented results accurately and comprehensively reflect how our proposed approach and existing baselines perform on both political think tanks. For text-based baselines, we encode Wikipedia summaries of entities with these methods and predict their stances with two fully connected layers. For graph-based baselines, we train them with the constructed HIN and the expert knowledge training objective.


\begin{table}[t]
    \centering
    \begin{tabular}{c|c|c|c}
         \toprule[1.5pt] 
         \textbf{Hyperparameter} & \textbf{Value} & \textbf{Hyperparameter} & \textbf{Value}\\
         \midrule[0.75pt]
         RoBERTa size & $768$ & GNN size & $512$ \\ 
         optimizer & Adam & learning rate & $1e-3$ \\
         batch size & $64$ & max epochs & $100$ \\
         $L$ & $2$ & $\phi$ & ReLU \\
         $Q$ & $-0.1$ & \#negative sample & $2$ \\
         $\lambda_1$ & $0.01$ & $\lambda_2$ & $0.2$ \\
         $\lambda_3$ & $1$ & $\lambda_4$ & $1e-5$ \\ \bottomrule[1.5pt]
    \end{tabular}
    \caption{Hyperparameters of our proposed approach.}
    \label{tab:hyperparameter}
\end{table}

\end{document}